\documentclass[letterpaper, 10 pt, conference]{ieeeconf}  

\IEEEoverridecommandlockouts                              
\title{\LARGE \bf
The Uncertainty Aware Salted Kalman Filter: State Estimation for Hybrid Systems with Uncertain Guards}

\author{J. Joe Payne, Nathan J. Kong, and Aaron M. Johnson
    \thanks{This material is based upon work supported by the U.S. Army Research
    Office under grant \#W911NF-19-1-0080 and the National Science Foundation
    under grant \#CMMI-1943900. The views and conclusions contained
    in this document are those of the authors and should not be interpreted as
    representing the official policies, either expressed or implied, of the Army Research Office, the National Science Foundation, or the U.S. Government.
    The U.S. Government is authorized to reproduce and distribute reprints
    for Government purposes notwithstanding any copyright notation herein.}%
    \thanks{All authors are with the Department of Mechanical Engineering, Carnegie Mellon University, Pittsburgh, PA, USA, \texttt{jjpayne@andrew.cmu.edu}}%
}
\usepackage{amsmath}
\usepackage{mathtools}
\usepackage{amsthm}
\usepackage{amssymb}
\usepackage{accents}
\usepackage{cite}

\usepackage{algorithm}
\usepackage{algpseudocode}
\usepackage[bookmarks=true]{hyperref}
\hypersetup{%
  breaklinks,
  bookmarksnumbered=true,
  bookmarksopen=true,
  colorlinks=true,
  linkcolor=black,
  urlcolor=black,
  citecolor=black
}

\newcounter{theorem}

\newtheorem{definition}[theorem]{Definition}

\begin{document}

\maketitle
\thispagestyle{empty}
\pagestyle{empty}

\begin{abstract}
In this paper, we present a method for updating robotic state belief through contact with uncertain surfaces and apply this update to a Kalman filter for more accurate state estimation.
Examining how guard surface uncertainty affects the time spent in each mode, we derive a novel guard saltation matrix -- which maps perturbations prior to hybrid events to perturbations after -- accounting for additional variation in the resulting state. 
Additionally, we propose the use of parameterized reset functions -- capturing how unknown parameters change how states are mapped from one mode to the next -- the Jacobian of which accounts for additional uncertainty in the resulting state.
The accuracy of these mappings is shown by simulating sampled distributions through uncertain transition events and comparing the resulting covariances. 
Finally, we integrate these additional terms into the ``uncertainty aware Salted Kalman Filter'', uaSKF, and show a peak reduction in average estimation error by 24-60\% on a variety of test conditions and systems.
\end{abstract}

\section{Introduction}
Making and breaking contact is critical for robots as they often need to physically interact with their environment to accomplish their tasks. 
For a legged robot to navigate to a desired location -- for search and rescue, mapping, remote surveying, etc. -- its feet will need to repeatedly impact the ground as it walks or runs. 
Manipulation robots must grasp, push, pull, etc, the objects they need to manipulate.
In order to safely and reliably operate during these  changing contact conditions, robots need to have an accurate estimation of their state in order to generate reasonable plans and complete their tasks.
However, when dealing with these intermittent contacts, the robot's dynamics become non-smooth and even discontinuous, which presents a challenge for classic methods that assume smoothness \cite{mitcheetahstateestimation2018,bloesch2013state, hartley2020contact, varin2018constrained}.

\begin{figure}[tb]
    \centering
    \includegraphics[width = \linewidth]{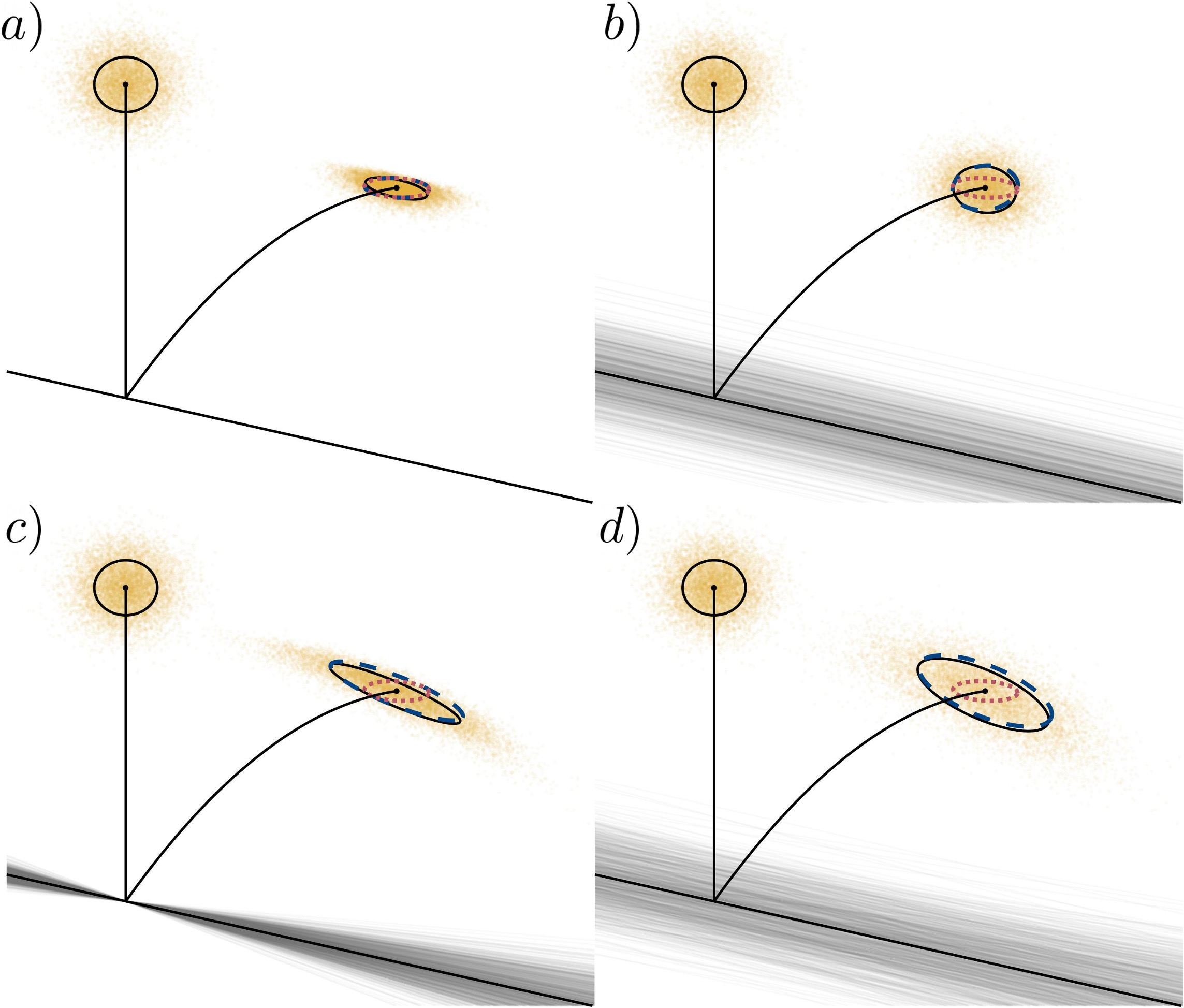}
    \caption{Simulating a 2D bouncing ball impacting an angled ground. a) No uncertainty in the guard offset or normal. b) Only offset (guard location) uncertainty. c) Only surface normal (reset function) uncertainty. d) Both offset and normal uncertainty.
    Yellow particles: Initial and final distribution. Black curve: Nominal trajectory. Black ellipses: Initial and final covariances. Red dotted ellipse: Predicted covariance using only saltation matrix (identical in all four plots). Blue dashed ellipse: Predicted covariance using the proposed method. The proposed method captures the effect of guard and reset uncertainty on the propagated covariance.
    }
    \label{fig:guard_normal_uncertainty_propagation}
\end{figure}

Another difficulty with intermittent contact systems is that outside of constrained environments like labs and factories, there will not be perfect models of the environment. The contact surface and physical properties may not be perfectly known ahead of time.
In the language of hybrid systems
\cite{Back_Guckenheimer_Myers_1993,goebel2009hybrid,LygerosJohansson2003}, this environmental uncertainty requires us to consider the guards (where contact conditions change) and reset maps (how contact conditions change) to be stochastic.
The combination of uncertainty in the guard with the discontinuity in dynamics results in additional state uncertainty due to variation in the time spent in each mode.
For example, Fig.~\ref{fig:guard_normal_uncertainty_propagation} shows a simple contact system -- a ball bouncing off of a slanted surface. If there is uncertainty in the guard location or in the angle at which the ball will rebound, the resulting state uncertainty will grow. 
Filtering methods present a way to utilize the information of how state and guard uncertainty interact at hybrid events by directly updating the state belief based on hybrid model uncertainty.

To address these issues, this paper presents an ``uncertainty aware Salted Kalman Filter'', or uaSKF, for hybrid dynamical systems.
We propose to model uncertainty as distributions of guard locations and reset map parameters, and we derive how these distributions couple into the system state uncertainty.
We introduce the ``guard saltation matrix,'' which captures the uncertainty due to variations in the guard location and time to impact (Fig.~\ref{fig:guard_normal_uncertainty_propagation}b). 
To handle reset uncertainty, we use the Jacobian of the reset function with respect to the uncertain parameters (Fig.~\ref{fig:guard_normal_uncertainty_propagation}c). 
These terms combine to produce an accurate state uncertainty update through the uncertain ground interaction (Fig.~\ref{fig:guard_normal_uncertainty_propagation}d).
Then we integrate these tools into a Kalman filter and provide results showing reduced estimation error on several example systems.

\section{Related Work}
While the fields of hybrid systems and state estimation both have long histories, here we focus on related work that specifically tackles the intersection of the two.

\subsection{State Estimation for Hybrid Systems}
Initial work on Kalman filter (KF) based methods for hybrid dynamical systems such as \cite{bloesch2013state, hartley2020contact} utilized the reset map to update the mean estimate and the Jacobian of the reset map to update covariance beliefs at hybrid events.
However, recent work showed that the Jacobian does not capture all of the effects of the hybrid event on reshaping the covariance. 
Instead, \cite{kong2021salt} proposes the ``Salted Kalman Filter'' (SKF), which uses the saltation matrix in place of the Jacobian of the reset map for Kalman filtering. The saltation matrix is a rank 1 update to the Jacobian that accounts for state variations caused by time to impact variations. 
This work was extended to include filtering on manifolds in \cite{GAO2021290} with the hybrid invariant extended Kalman filter (HInEKF).
In this work, the saltation matrix method is further extended to account for uncertainties in the structure of the underlying hybrid system, such as variation in guard location and the reset map.

Other work on hybrid system state estimation has largely involved multiple estimators. 
Some have kept the number of estimators relatively low, such as in Interacting Multiple Model estimation (IMM) \cite{blom1988interacting}, which maintains KFs for each of the hybrid modes.
Multiple model methods have been extended to a variety of problems including nonlinear dynamics \cite{barhoumi2012observer} and non-identity rests \cite{BALLUCHI2013915}.
Multiple model methods are not easily applicable on event-driven hybrid dynamical systems as one of the core assumptions of these methods is that the transitions between discrete modes follow a Markov model, which is not necessarily true when the probabilities of discrete state transitions are dependent on the continuous state beliefs.

Alternatively, many methods such as \cite{koutsoukos2002monitoring,koval2017manifold} have adopted particle filtering approaches and have used large numbers of individual estimates to represent a distribution, as opposed to summary statistics like mean and covariance in the case of KFs. 
While these methods have many benefits, including capturing nonlinear dynamics and non-Gaussian beliefs, the computational complexity of running a particle filter is far greater than a simpler Kalman style filter. 
As such, this work aims to utilize KFs to maintain the benefits of fast computation times.

\subsection{Nonlinear Event Mapping and Saltation Matrices}
In this paper, the standard KF is augmented with additional knowledge about the structure of reset maps from the saltation matrix.

The saltation matrix \cite{AIZERMAN19581065,hiskens2000trajectory,leine2013dynamics,burden2016event} is used to map perturbations through nonsmooth dynamics at the boundary between modes. 
Previously, \cite{biggio2014accurate} demonstrated the saltation matrix can be used to map probability distributions through hybrid transitions and \cite{kong2021salt} extended this to use in Kalman filtering, as in (Fig.~\ref{fig:guard_normal_uncertainty_propagation}a).

The primary difference in this work is that perfect knowledge of guard locations and reset maps is not assumed. 
This is similar to \cite{staunton2020noisy} in which they examined the ``noisy saltation matrix'' for systems with guards that are time varying with random low amplitude, zero mean, mean reverting noise. 
This paper uses similar time to impact analysis to derive our guard saltation matrix, which instead views the uncertainty in guard locations as stationary with an estimated distribution they are drawn from. 
This results in a different mapping, as the noisy saltation matrix views the guard as time varying (so the velocity of the noise affects the time to impact for the system), which is not present in our guard saltation matrix. 
The other difference is that the noisy saltation matrix does not assume any sort of distribution, only that information about time to impact can be extracted. 
This work assumes that Gaussian information is known about the guard and directly determines the time to impact from that distribution.

\section{Background}

\subsection{Hybrid Dynamical Systems}
A hybrid dynamical system is a system with both continuous states, such as positions and velocities, and discrete states or modes, such as whether a specific limb is in contact with the ground, in which the sequence of discrete states is determined by the evolution of the continuous states. 
More formally, closely following \cite[Def.~2]{johnson2016hybrid}:
\begin{definition} \label{def:hs}
    A $C^r$ \textbf{hybrid dynamical system}, for continuity class $r\in \mathbb{N}_{>0} \cup \{\infty,\omega \}$, is a tuple $\mathcal{H} := (\mathcal{J},{\mathnormal{\Gamma}},\mathcal{D},\mathcal{F},\mathcal{G},\mathcal{R})$ whose constituent parts are defined as:
    \begin{enumerate}
        \item $\mathcal{J} := \{I,J,...,K\} \subset \mathbb{N}$ is the set of discrete \textbf{modes}.
        \item $\mathnormal{\Gamma} \subset \mathcal{J}\times\mathcal{J}$ is the set of discrete \textbf{transitions} forming a directed graph structure over $\mathcal{J}$.
        \item $\mathcal{D}:=\amalg_{{I}\in\mathcal{J}}$ ${D}_{I}$ is the collection of \textbf{domains},
        \item $\mathcal{F}:= \amalg_{I\in\mathcal{J}} F_I$ is a collection of $C^r$ time-varying \textbf{vector fields}, $F_I:  \mathbb{R}\times D_I\to\mathcal{T}D_I$.
        \item $\mathcal{G}:=\amalg_{(I,J)\in\mathnormal{\Gamma}}$ $G_{I,J}(t)$ is the collection of \textbf{guards}, where $G_{I,J}(t)\subset D_I$ for each $(I,J)\in \mathnormal{\Gamma}$ is defined as a sublevel set of a $C^r$ function, i.e.\ $G_{I,J}(t)= \{x \in D_I|g_{I,J}(t,x)\leq0\}$.
        \item $\mathcal{R}:\mathbb{R}\times \mathcal{G}\rightarrow \mathcal{D}$ is a $C^r$ map called the \textbf{reset} that restricts as $R_{I,J}:=\mathcal{R}|_{G_{I,J}(t)}:G_{I,J}(t)\rightarrow D_J$ for each $(I,J)\in \mathnormal{\Gamma}$.
    \end{enumerate}
\end{definition}

\subsection{Perturbation Analysis and the Saltation Matrix}
Within a single mode, the Jacobian of the continuous dynamics can be used to update the covariance of a distribution. 
However, at hybrid events, the same method cannot be used. 
In order to properly update covariance through a mode transition, the time to impact of perturbations must be considered. 
The saltation matrix includes the time to impact for covariance updates \cite{AIZERMAN19581065,hiskens2000trajectory,leine2013dynamics,burden2016event}:
\begin{align}
\Xi_x := D_xR_{I,J} + \frac{(f_J-D_xR_{I,J}f_I - D_tR_{I,J})D_xg_{I,J}}{D_xg_{I,J}f_I + D_tg_{I,J}}
\label{eq:saltation}
\end{align}
where $D_x$ and $D_t$ represent Jacobians with respect to state and time, $f_I$ is the linearization of the vector field $F_I$ at the point of impact.
The saltation matrix captures both the effects of the Jacobian of the reset map and variations in the time a system is acted upon by the dynamics of each mode. It maps pre-transition variations $\delta x^-$ to post-transition variations $\delta x^+$ as, $\delta x^+ = \Xi_x \delta x^-$,
and, by extension \cite{kong2021salt,biggio2014accurate}, maps pre-transition covariance in state $\Sigma_x^-$ to post-transition covariance $\Sigma_x^+$ as:
\begin{align}
    \Sigma_x^+ = \Xi_x \Sigma_x^- \Xi_x^T
    \label{eq:saltcov}
\end{align}

In order to ensure the saltation matrix is well defined, we utilize the conventional assumptions from \cite[Assumption 1]{burden2018contraction}, which most notably includes that transitions are transverse.
This ensures that trajectories in a neighborhood of a guard must transition exactly once at small timescales. This assumption excludes Zeno behavior from this analysis.
Additionally, we make the assumption that the vector fields in each mode are extensible beyond the nominal guard as is done in \cite{staunton2020noisy}.
This assumption allows us to analyze the effect of each mode's dynamics on a trajectory through a guard that is not at the nominal position.

\section{Modeling Uncertainty in Guards and Reset Maps}
A notable limitation of the saltation matrix formulation is that it assumes perfect knowledge of the structure of the hybrid system.
However, in real applications, the guard boundaries and reset properties will be uncertain. 
This work seeks to capture these uncertainties for covariance propagation in hybrid systems with uncertainty in the guard (Sec.~\ref{sec:uncertain_guard}) and reset (Sec.~\ref{sec:uncertain_reset}).

\begin{figure}[t]
    \centering
    \includegraphics[width = \linewidth]{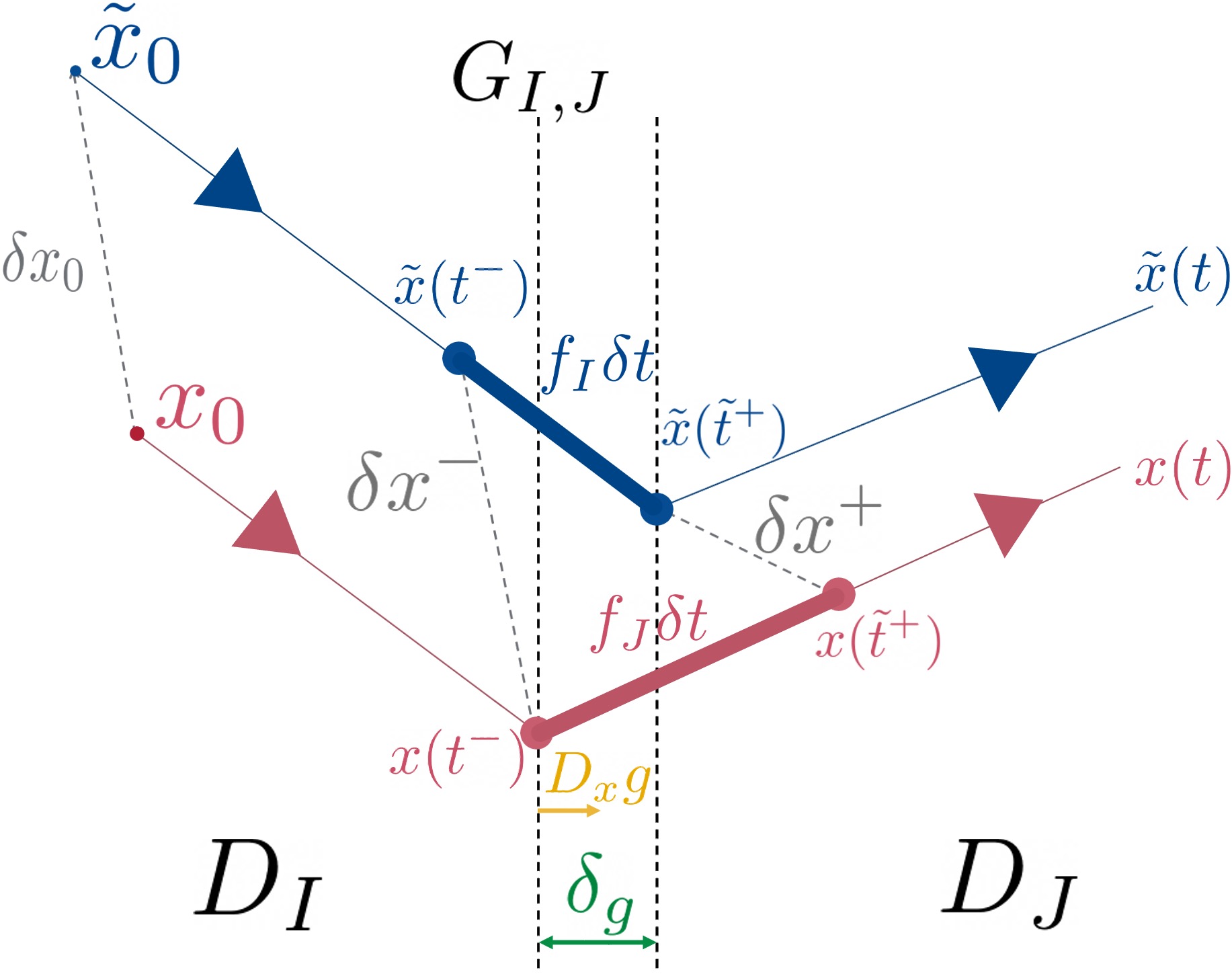}
    \caption{Uncertainty in guard location changing the relative position of variations before and after transition. For ease of depiction, the reset map is shown as identity, but accounting for them is an additive term in the saltation matrix.
    This is the basis of the guard saltation matrix.}
    \label{fig:uncertainGuardMath}
\end{figure}

\subsection{Uncertainty in Guard Location}
\label{sec:uncertain_guard}
When the location of the guard is unknown along its normal direction, the post-transition state uncertainty is higher than if the guard location is perfectly known. 
The effects of uncertainty along the guard normal direction can be captured as an additional rank 1 update to the standard saltation matrix.
Note that the effects of uncertainty in the normal direction of the guard surface are discussed in (Sec.~\ref{sec:uncertain_reset}), as the timing elements of uncertainty in normal are higher order terms that do not show up for a first order approximation, similar to the curvature of the guard surface.
This section goes through the mathematical derivation of the guard-uncertainty saltation matrix, using a geometric derivation similar to the saltation matrix derivation in \cite{leine2013dynamics}.

Here we consider how pre-impact displacements $\delta x(t^-)$ in mode $I$ map to post-impact displacements $\delta x(\tilde{t}^+)$ in mode $J$.
To do so, we start by considering how the nominal state $x(t)$ and a perturbed state $\tilde{x}(t)$ evolve from the time the nominal state reaches the guard before transitioning $(t^-)$ to the time the perturbed state enters the new mode $(\tilde{t}^+)$. For simplicity of notation, in this derivation we assume that the nominal trajectory reaches the guard first, however, the same result is reached if the opposite assumption is chosen. The displacements can be expanded as:
\begin{align}
\delta x(t^-) & = \tilde{x}(t^-) - x(t^-) \\
\delta x(\tilde{t}^+) & = \tilde{x}(\tilde{t}^+) - x(\tilde{t}^+) \label{eq:delta_x_plus}
\end{align}

For readability, hereafter we use the + and - superscripts to represent the state at times $\tilde{t}^+$ and $t^-$, respectively, e.g.\ $x^+= x(\tilde{t}^+)$.
The evolution of these trajectories can be seen in Fig.~\ref{fig:uncertainGuardMath}.

We would like to solve for $\delta x^+$ as a function of $\delta x^-$ and known system parameters by using the continuous and reset dynamics. For this first-order analysis we linearize the pre-impact dynamics $F_I$ about the point $x(t^-)$ as $f_I$, and similarly we linearize the post-impact dynamics $F_J$ about the point $x(t^+)$ as $f_J$.
Following the linearized hybrid dynamics forward from time $t^-$, and assuming without loss of generality that the nominal trajectory impacts the guard first at time $t^-$, we have:
\begin{align}
    x^+ = R_{I,J}(x^-,t^-) + f_J \delta t
    \label{eq:x_plus}
\end{align}
that is, the final state is equal to the initial state passed through the reset map and then following the dynamics of the new mode for time $\delta t=\tilde{t}-t$, the time between impact events, until time $\tilde{t}^+$. 
Similarly, we can follow the hybrid dynamics forward from $\tilde{x}^-$ to get:
\begin{align}
    \tilde{x}^+ = R_{I,J}(\tilde{x}^- + f_I\delta t,\tilde{t}^-)
    \label{eq:tilde_x_plus}
\end{align}
where in this case the perturbed state first flows in the prior mode until time $\tilde{t}^-$ and then passes through the reset. 
Using the definition of $\delta x^-$ and $\delta t$, \eqref{eq:tilde_x_plus} can be written entirely in terms of the pre-impact variation and times:
\begin{equation}
\begin{split}
\tilde{x}^+ & = R_{I,J}(x^- + \delta x^- + f_I\delta t,  t^- + \delta t) 
\label{eq:tilde_x_plus_exp}
\end{split}
\end{equation}
Substituting \eqref{eq:x_plus} and \eqref{eq:tilde_x_plus_exp} back into \eqref{eq:delta_x_plus} yields:
\begin{align}
\delta x^+ =&  R_{I,J}(x^- + \delta x^- + f_I\delta t, t^- + \delta t) \nonumber\\
& - R_{I,J}(x^-,t^-) - f_J \delta t
\end{align}

Using the first-order Taylor series expansion of the reset map about $(x^-,t^-)$, we can replace this expression with:
\begin{align}
\delta x^+ =&  R_{I,J}(x^-,t^-) + D_xR_{I,J}\delta x^- \nonumber
 + D_xR_{I,J}f_I\delta t \\
 & + D_tR_{I,J} \delta t 
 - R_{I,J}(x^-,t^-) - f_J \delta t \\
 =& D_xR_{I,J}\delta x^- 
+ \left( D_xR_{I,J}f_I + D_t R_{I,J} - f_J \right) \delta t
\label{eq:dxplus}
\end{align}

The next step is to determine what $\delta t$ is in terms of $\delta x^-$ and system parameters by examining the dynamics of the first mode along the guard normal direction, $D_xg$:
\begin{align}
(D_xgf_I + D_t g)\delta t = -D_xg\delta x^- + \delta_g \\
\delta t = \frac{-D_xg\delta x^- + \delta_g}{D_xgf_I + D_tg}
\label{eq:deltat}
\end{align}
where $\delta_g$ is the perturbation in the guard location along its normal direction, which holds whether the guard occurs early or late. 
Effectively, this means that time is equal to distance divided by velocity on infinitesimal perturbations. 
Note that this additional perturbation $\delta_g$ is the key difference compared to the derivation of the traditional saltation matrix.
Additionally, it should be noted that this requires the extra assumption that the dynamics are extensible beyond the nominal guard surfaces.

Plugging \eqref{eq:deltat} into \eqref{eq:dxplus} results in the following expression:

\begin{align}
\delta x^+ =& \left(D_xR_{I,J} + \frac{(f_J-D_xR_{I,J}f_I - D_tR_{I,J})D_xg}{D_xgf_I + D_tg}\right)\delta x^- \nonumber \\
& +\left(\frac{D_xR_{I,J}f_I + D_tR_{I,J}-f_J}{D_xgf_I + D_tg}\right)\delta_g\\
=& \Xi_x \delta x^- + \Xi_g \delta_g
\end{align}
where $\Xi_x$ is the traditional saltation matrix, \eqref{eq:saltation}, and $\Xi_g$ is a ``guard saltation matrix'', defined as:
\begin{align}
    \Xi_g := \frac{D_xR_{I,J}f_I + D_tR_{I,J} -f_J}{D_xgf_I + D_tg} \label{eq:guard_saltation}
\end{align}
Note that $\Xi_x = D_xR_{I,J} - \Xi_g D_xg$, and so $\Xi_g$ can be computed as part of computing $\Xi_x$. The guard saltation matrix is a single column vector if $\delta_g$ is only the normal direction component of guard uncertainty\footnote{For a full dimensional guard uncertainty vector $\delta g^-$, we have $\delta_g = D_xg \delta g^-$, and can use the matrix $(\Xi_g D_xg)$ as $\delta x^+ = \Xi_x \delta x^- + (\Xi_g D_xg) \delta g^-$, however note that only the uncertainty along the normal direction affects the outcome.}. Note also that while \eqref{eq:guard_saltation} was derived assuming the nominal transitions first, the same expression is obtained if the perturbed trajectory is assumed to transition first.

This can be used as an extended saltation matrix, which we call $\hat{\Xi}$, 
\begin{equation}
\begin{split}
\begin{bmatrix} \delta x^+ \\ \delta_g \end{bmatrix} = \begin{bmatrix}
\Xi_x & \Xi_g \\
0 & 1
\end{bmatrix} \begin{bmatrix} \delta x^- \\ \delta_g \end{bmatrix} = \hat{\Xi} \begin{bmatrix} \delta x^- \\ \delta_g \end{bmatrix}
\end{split}
\end{equation}

Extending now to covariance, as in \eqref{eq:saltcov}, when using this saltation matrix with no prior known covariance between the guard and state, the covariance updates for both the state and the guard are:
\begin{equation}
\begin{split}
\begin{bmatrix}
\Sigma_x^+ & \Sigma_{xg}^+ \\
\Sigma_{gx}^+ & \Sigma_g^+
\end{bmatrix} = 
\hat{\Xi}
\begin{bmatrix}
\Sigma_x^- & 0 \\
0 & \Sigma_g^-
\end{bmatrix}
\hat{\Xi}^T
\label{eq:extended_sigma}
\end{split}
\end{equation}
where $\Sigma_x$ is the state covariance and $\Sigma_g$ is the guard covariance. 
Pulling out the state covariance through an uncertain guard, we get:
\begin{equation}
\begin{split}
\Sigma_x^+ = \Xi_x\Sigma_x^-\Xi_x^T + \Xi_g\Sigma_g^-\Xi_g^T
\label{eq:sigma_update}
\end{split}
\end{equation}

The improved covariance estimation of the guard saltation matrix is shown in Fig.~\ref{fig:guard_normal_uncertainty_propagation}b. 
Note that the covariance of the uncertain guard distribution is equivalent to the sum of the covariance of propagating the initial distribution through the certain guard and the resulting covariance of propagating a known starting condition through an uncertain guard.
For this trial, the K-L divergence \cite{KullbackLeiblerDiv} (which is a measure of the difference between two probability distributions) between the actual covariance and the estimated is reduced from 402 to 0.03 by including the guard uncertainty propagation terms.
The final error between the true and the estimated covariance from the guard saltation matrix is caused by the linearization of the dynamics.

There are trade-offs between using the extended matrix \eqref{eq:extended_sigma} and directly updating the state covariance \eqref{eq:sigma_update}. 
Using the extended matrix requires extending the state and increases the dimensionality of all terms, but it allows for past measurements to affect knowledge of the guard distribution.
Directly updating the state distribution allows for a simpler state vector, but assumes that the guard distribution is static.
If repeated behavior near the same region of a guard is expected, then it would be beneficial to try to estimate the guard parameters. 
However, in situations like locomotion on uneven ground where the system is not expected to re-traverse the same areas frequently, it makes sense to accept the mean and covariance as fixed (or calculated separately, e.g.\ based on exteroceptive sensor noise) parameters for the guard surface.

\subsection{Uncertainty in Reset Parameters}
\label{sec:uncertain_reset}
Even in cases where the location of the guard is perfectly known, the exact properties of the reset map may not be known.
Some physical examples of this include the coefficient of restitution in elastic systems and the precise surface normal in any contact system.
These types of uncertainties can be handled by parameterizing the reset maps to be not only functions of state and time, but to also include other parameters.
By including these additional parameters, the resulting uncertainty in state caused by variations in reset parameters can be examined through the Jacobian with respect to these additional parameters.

For this derivation, we re-define the reset map $R(x)$ to include its ``fixed'' parameters as arguments $R(x,\theta)$. 
This $\theta$ term can include values like the coefficient of restitution in the bouncing ball problem. 
Using this formulation, we can examine how uncertainty in model parameters can affect the resulting state estimation covariance after an impact,
updating \eqref{eq:x_plus} and \eqref{eq:tilde_x_plus}:
\begin{align}
x^+ & = R_{I,J}(x^-,\theta) + f_J\delta t \\
\tilde{x}^+ &= R_{I,J}(\tilde{x}^- + f_I\delta t, \tilde{\theta})
\end{align}
Taking the difference between these two to find $\delta x^+$:
\begin{equation}
\begin{split}
\delta x^+ = R_{I,J}(\tilde{x}^- + f_I\delta t, \tilde{\theta}) - (R_{I,J}(x^-,\theta) + f_J\delta t)
\end{split}
\end{equation}

Now using first order approximations of the reset map with the Taylor series expansion, as in \eqref{eq:dxplus}:
\begin{align}
\delta x^+ =&  R_{I,J}(x^-,\theta) + D_xR_{I,J}\delta x^-+ D_xR_{I,J}f_I\delta t  \nonumber\\ 
& + D_{\theta}R_{I,J}\delta\theta + D_tR_{I,J}(x^-,\theta)\delta t \nonumber\\
&-R_{I,J}(x^-,\theta)-f_J\delta t\\
=& D_xR_{I,J}\delta x^- + \left(D_xR_{I,J}f_I + D_t R_{I,J} - f_J\right)\delta t \nonumber\\
&+ D_{\theta}R_{I,J}\delta\theta
\end{align}
where equality holds to first order.

By using $\delta t$ from \eqref{eq:deltat}, and rearranging into a block matrix, this becomes:
\begin{equation}
\begin{split}
\begin{bmatrix}
\delta x^+ \\
\delta \theta^+ 
\end{bmatrix}
=
\begin{bmatrix}
\Xi_x & D_{\theta}R_{I,J} \\
0 & 1
\end{bmatrix}
\begin{bmatrix}
\delta x^- \\
\delta \theta^- 
\end{bmatrix}
\end{split}
\end{equation}

Assuming there is no initial covariance between the state and the reset map information, this can be used to update the state covariance with:
\begin{equation}
\begin{split}
\Sigma_x^+ = \Xi_x\Sigma_x^- \Xi_x^T + (D_{\theta}R_{I,J})\Sigma_{\theta}^- (D_{\theta}R_{I,J})^T
\end{split}
\end{equation}
where $\Sigma_{\theta}^-$ is the covariance of the $\theta$ parameters. Note that, as was the case with uncertain guard, the covariance of the uncertain reset distribution is equivalent to the sum of the covariance of propagating the initial distribution through the certain reset and the resulting covariance of propagating a known starting condition through an uncertain reset.

This expression can be used in combination with the uncertainty in guard location for a total covariance update:
\begin{align}
\Sigma_x^+ = \Xi_x\Sigma_x^- \Xi_x^T 
+ \Xi_g\Sigma_g^-\Xi_g^T
+(D_{\theta}R)\Sigma_{\theta} (D_{\theta}R)^T
\label{eq:sigma_combined}
\end{align}

Results demonstrating the improvement over assuming perfect knowledge of the reset map in uncertainty propagation can be found in Fig.~\ref{fig:guard_normal_uncertainty_propagation}c. 
For this trial, the K-L divergence between the actual covariance and the estimated is reduced from 357.6 to 19.8 by including the reset uncertainty propagation term.
Furthermore, combining both guard and reset uncertainty, Fig.~\ref{fig:guard_normal_uncertainty_propagation}d, the K-L divergence between the actual covariance and the estimated is reduced from 739 to 0.03.
The final error between the true and the estimated covariance is caused by the linearization of the dynamics.

\section{Kalman Filtering with Uncertain Environment}
This section applies the covariance update rules found in the prior section to the problem of state estimation using Kalman filtering (summarized briefly in Sec.~\ref{sec:kf}).
With these more accurate distribution updates through hybrid events, we can achieve better estimation accuracy.
The implementation details for this ``uncertainty aware SKF'' (uaSKF) follow the algorithm for SKF, presented in \cite{kong2021salt}.
The key differences from the SKF occur during hybrid transition events.
We discuss how to handle these events in both the process and measurement updates in Sec.~\ref{sec:apriori} and Sec.~\ref{sec:aposteriori}, respectively.

\subsection{Kalman Filtering in the Smooth Domains}
\label{sec:kf}
While the system is not interacting with any guard surfaces, the uaSKF behaves as a standard KF or EKF would.
In these cases, the system follows the standard update rules.
The standard KF updates the mean $\hat{x}$ and covariance $\hat{\Sigma}$ estimate in two steps  \cite[Eqns.~1.9--1.13]{welch1995introduction}:
first, the \emph{a priori} update:
\begin{align}
    \hat{x}(k+1|k) &= A_{I,\Delta}\hat{x}(k) \label{eq:apriorix}\\
    \hat{\Sigma}(k+1|k) &= A_{I,\Delta}\hat{\Sigma}(k)A_{I,\Delta}^T + W_{I,\Delta} \label{eq:apriorisigma}
\end{align}
at timestep $k+1$, where $A_{I,\Delta}$ is the discrete dynamic matrix for $F_I$, $\Delta$ is the discretization timestep, and $W_{I,\Delta}$ is the covariance of the additive Gaussian process noise. Second, the \emph{a posteriori} update:
\begin{align}
    &K_{k+1} = \hat{\Sigma}(k+1|k)C_I^T\left[C_I\hat{\Sigma}(k+1|k)C_I^T + V_I\right]^{-1}\!
    \label{eq::kalmangain}\\
    &\hat{x}(k+1|k+1) = \hat{x}(k+1|k) \label{eq::postmean}\\ &\qquad \qquad \qquad \quad+ K_{k+1}\left[y(k+1) - C_I\hat{x}(k+1|k)\right] \nonumber\\
    &\hat{\Sigma}(k+1|k+1) = \hat{\Sigma}(k+1|k) - K_{k+1}C_I\hat{\Sigma}(k+1|k)
    \label{eq::postcov}
\end{align}
where $K_{k+1}$ is the Kalman gain, $C_I$ is the measurement function, $y(k+1)$ is the measurement, and $V_I$ is the covariance of the additive Gaussian measurement noise.

While the standard Kalman update works well during the continuous domains, additional consideration has to be taken for hybrid events.
We will now present how to handle hybrid transitions in both the \emph{a priori} and \emph{a posteriori} updates.

\subsection{Uncertainty Aware Hybrid A Priori Updates}
\label{sec:apriori}
In a discretized KF, the hybrid event will most likely not take place perfectly at the time steps.
Accordingly, the \emph{a priori} update must include the first mode dynamics, the discrete update, and the second mode dynamics to bridge the time from one sample time to the next.
Conceptually, what happens here is three separate updates (or more if multiple hybrid transitions occur in a single timestep) combined into one, with the first and final portions following \eqref{eq:apriorix}--\eqref{eq:apriorisigma} with timesteps $\Delta_1$ and $\Delta_2$, respectively.
This does not require knowledge of the number of impact events in the system as these updates are driven by the mean estimate reaching guards, not by ground truth impacts.

The instantaneous mean and covariance updates during the discrete transition from mode $I$ to $J$ use the following update rules, using \eqref{eq:sigma_combined} as in \cite[Eqns.~16--17]{kong2021salt}\footnote{In the notation of \cite[Eqn.~17]{kong2021salt}, we are setting\\
$W_{R_{(I,J)}} = \Xi_g\Sigma_g\Xi_g^T + D_\theta R \Sigma_\theta D_\theta R^T$.}:
\begin{align}
    x_J(t) &= R_{I,J}(x_I(t))
    \label{eq::kfresetdynamicupdate}\\
    \Sigma_J(t) &=\Xi_x\Sigma_I(t)\Xi_x^T
    \label{eq::kfcovdynamicupdate}
    +\Xi_g\Sigma_g\Xi_g^T + (D_\theta R_{I,J}) \Sigma_\theta (D_\theta R_{I,J})^T
\end{align}

By combining this instantaneous update with the standard KF updates \eqref{eq:apriorix}--\eqref{eq:apriorisigma} in the first and second modes, as in \cite[Eqns.~18--19]{kong2021salt}, we arrive at the following update rule:
\begin{align}
    \hat{x}(k+1|k) =& A_{J,\Delta_2}R_{I,J}\big(A_{I,\Delta_1}\hat{x}(k)\big)
    \label{eq::kf_mean_apriori}\\
    \hat{\Sigma}(k+1|k) =&  A_{J,\Delta_2}\Big(\Xi_x(A_{I,\Delta_1}\Sigma(k)A_{I,\Delta_1}^T+W_{I,\Delta_1})\Xi_x^T
    \nonumber\\
    &\!
    +\Xi_g\Sigma_g\Xi_g^T + D_\theta R \Sigma_\theta D_\theta R^T\Big)A_{J,\Delta_2}^T + W_{J,\Delta_2}
    \label{eq::covariance_dynamics}
\end{align}

\subsection{Uncertainty Aware Hybrid A Posteriori Updates}
\label{sec:aposteriori}
When a measurement update causes the mean estimate to meet a guard condition, the uncertainty aware SKF applies the discrete update to the a posteriori mean and covariance estimates following equations \eqref{eq::kfresetdynamicupdate}--\eqref{eq::kfcovdynamicupdate}.

\section{Kalman Filtering Performance on Example Systems}
To demonstrate the increased accuracy of these methods on systems with uncertainty in the guard conditions and reset dynamics, we provide filtering results from a variety of systems.
For each system, a comparison is made against the standard SKF formulation without incorporating knowledge of guard uncertainty using the sign test for the median difference between trials \cite{dixon1946statistical}, as used in \cite{kong2021salt}.

\begin{figure}[tb]
    \centering
    \includegraphics[width = 0.96\linewidth]{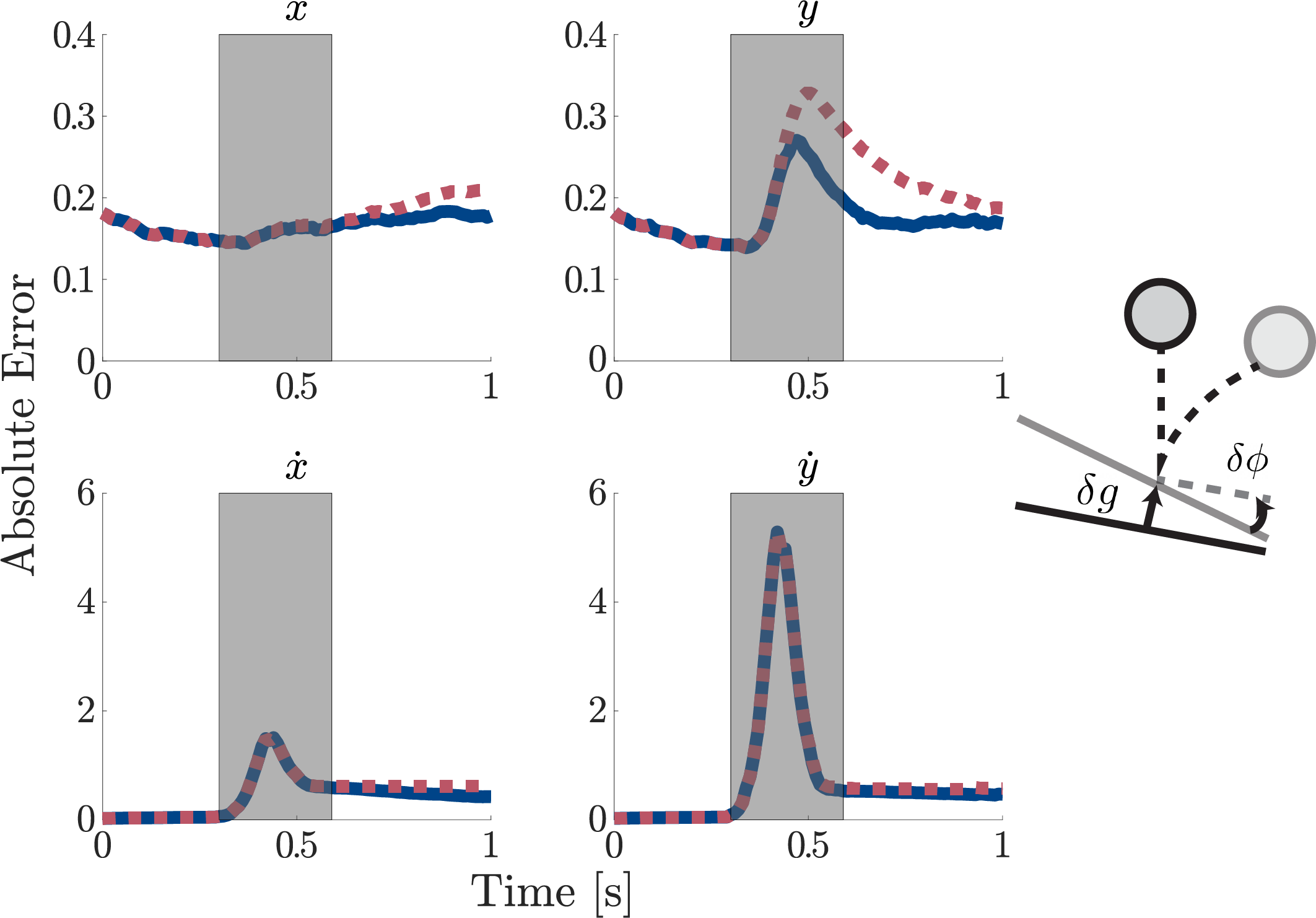}
    \caption{Average error for 1000 trials of Kalman filtering on a bouncing ball model with uncertain parameters. The standard SKF results are in red dashed lines and the uncertainty aware SKF results are in blue solid lines. The shading represents the time from the earliest to latest ground truth impacts. A sketch of the system and a typical trajectory is shown to the right.}
    \label{fig:kalman_filtering_error_comparison_2dball}
\end{figure}

\subsection{Elastic Bouncing Ball}
Consider a simple point mass in two dimensions $(x_1,x_2)$, elastically impacting a plane with uncertain height $\delta_g$, angle $\theta$, and coefficient of restitution $\alpha$. An example of propagating a distribution through this system can be seen in Fig.~\ref{fig:guard_normal_uncertainty_propagation}.
This system has 1 mode with a self reset.
The continuous dynamics are:
\begin{align}
    \dot{x} = \begin{bmatrix}
    \dot{x}_1,\dot{x}_2,\dot{x}_3,\dot{x}_4
    \end{bmatrix}^T = \begin{bmatrix}
    x_3,x_4,0,-a_g
    \end{bmatrix}^T
\end{align}
where $a_g$ is the acceleration due to gravity.

The guard for this system is:
\begin{align}
    g(x) = x_2\cos{\theta} - x_1\sin{\theta} - \delta_g
\end{align}

The reset dynamics for the system are:
\begin{align}
    x^+ = \begin{bmatrix}
    x_1^- \\
    x_2^- \\
    x_3^- + \sin{\theta} (1+ \alpha) (x_4^- \cos{\theta} - x_3^-\sin{\theta}) \\
    x_4^- - \cos{\theta} (1+ \alpha) (x_4^- \cos{\theta} - x_3^-\sin{\theta}) \\
    \end{bmatrix}
    \label{eq:elasticreset}
\end{align}

We simulated 1000 trials of this system running for 1 second from a height of 3 m with initial velocity of -5 m/s in the vertical direction and initial covariances of $0.05I_2$ for position and $0.001I_2$ for velocity, where $I_2$ is the $2\times 2$ identity matrix. For this test, the mean ground height was 0 with a standard deviation of 0.25 m, the mean angle of the ground was -0.25 radians with a standard deviation of 0.05 radians, and the coefficient of restitution was $0.8$. The process noise for this system was $W_p = 10I_2$ for position and $W_v = I_2$ for velocities. The system measured only positions with covariance $V = I_2$. 

The results of the elastic ball test are shown in Fig.~\ref{fig:kalman_filtering_error_comparison_2dball}. The uncertainty aware method had a lower error than the standard saltation method with $p < 0.005$.
The median improvement in MSE over complete 1 s trajectories is 0.6\%. Additionally, the maximum percent improvement in average error magnitude for a timestep is 24\% at $0.98$s. The estimation is improved in all dimensions, but most notably in the vertical direction because the uncertain guard component accounts for the extra uncertainty in the vertical direction, while the uncertain reset term arising from the uncertain guard normal adds uncertainty in both velocity terms, which propagates to position as the filter progresses.

\begin{figure}[tb]
    \centering
    \includegraphics[width = \linewidth]{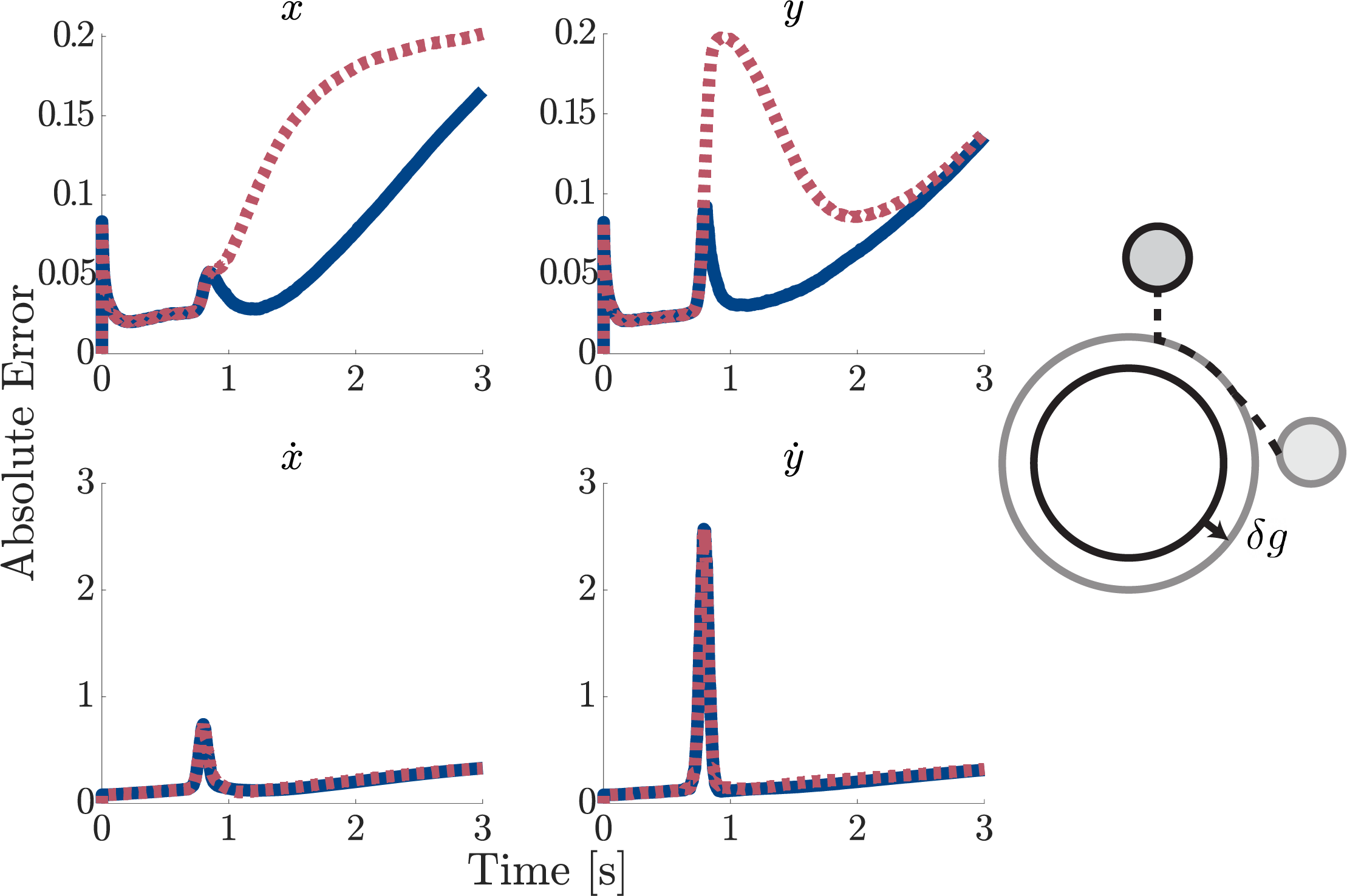}
    \caption{Average error of 1000 trials on a point mass plastically impacting a circle of unknown radius. The mean radius is 2m with a standard deviation of 0.25m. SKF results are red dashed lines and the uncertainty aware SKF results are blue solid lines. A sketch of the system and a typical trajectory is shown to the right. }
    \label{fig:kalman_filtering_error_comparison_circle_drop}
\end{figure}

\begin{figure*}[t]
    \centering
    \includegraphics[width = \textwidth]{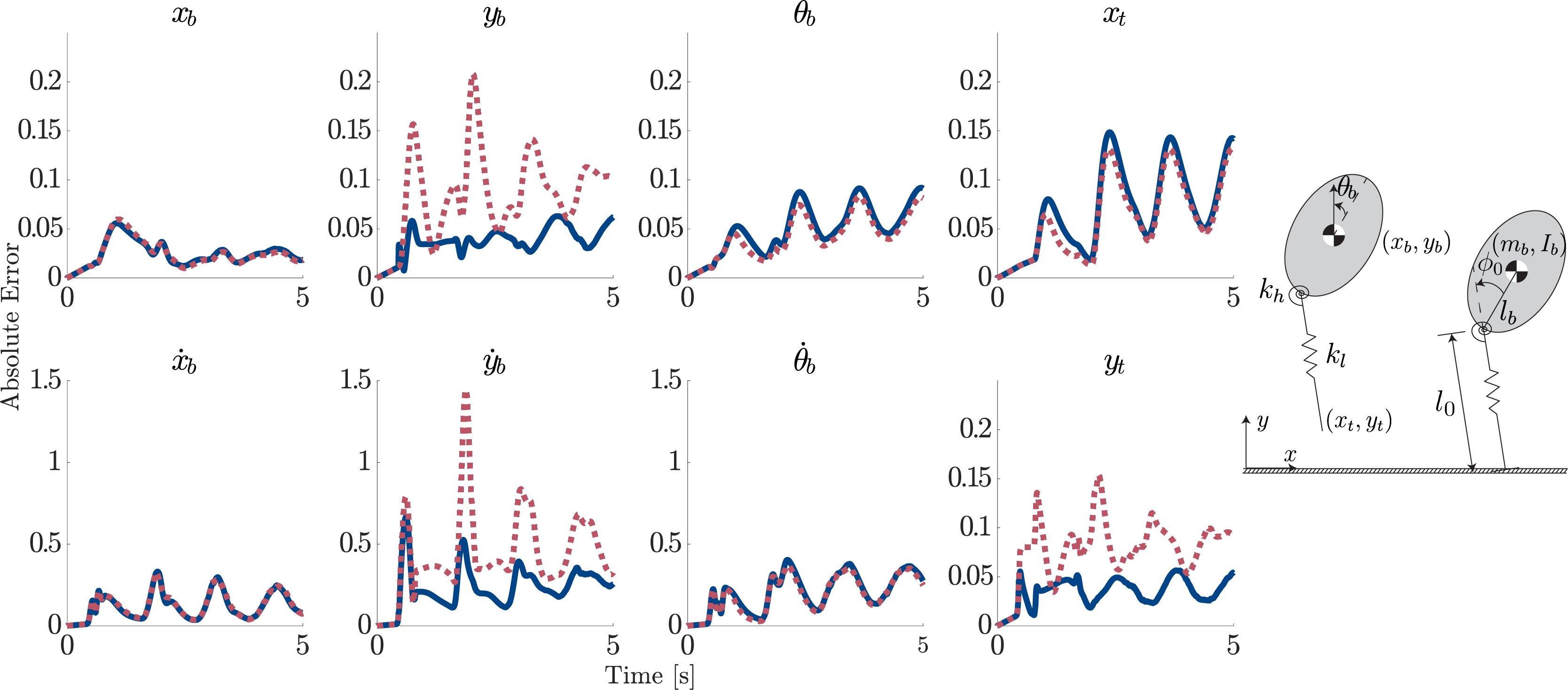}
    \caption{Left: Average error for 1000 trials on an ASLIP hopper with unknown ground height. The mean ground height was 0 with a standard deviation of 0.01m. The standard SKF results are in red dashed lines and the uncertainty aware SKF results are in blue solid lines. Note the improvement primarily in the vertical ($y$) directions that are aligned with the direction of ground uncertainty.  Right: The model used for these experiments.}
    \label{fig:kalman_filtering_error_comparison_ASLIP}
\end{figure*}

\subsection{Point Mass with Plastic Impact and Separation Conditions}
To demonstrate the method on a system with plastic impact and a constrained mode, we examine the case of dropping a plastic ball onto a circle of known center but unknown radius.
The system consists of two modes, an unconstrained aerial mode with the same dynamics as the elastic ball example, and a constrained mode when the ball is sliding on the surface.
The ballistic dynamics of this system are the same as the dynamics for the prior elastic ball example.
The impact reset map for this system is also the same as \eqref{eq:elasticreset}, however, the angle of the guard is determined by the tangent plane to the circle at the point of impact and the coefficient of restitution $\alpha$ is zero.
The constrained dynamics enforce zero acceleration along the radius extending from the center of the circle to the current position.
A liftoff event is triggered with an identity reset map when the force required to maintain the acceleration constraint becomes attractive.

We tested this system by running 1000 trials for 3 seconds each with initial position of (0.5m,5m), zero initial velocity, and position and velocity covariance of $0.1I_2$ each. The nominal surface is a circle with radius 2m centered about the origin.
The process noise for this system is $W_p = 0.1I_2$ for position and $W_v = 0.01I_2$ for velocity. 
The system measures positions with covariance $V = 0.1I_2$.

Results for this test can be seen in Fig.~\ref{fig:kalman_filtering_error_comparison_circle_drop}.
On average, utilizing awareness of model uncertainties notably improves performance on position estimates, but since the dynamics are largely the same regardless of minor position variations, the velocity estimates are very similar.
The uncertainty aware method outperforms the standard saltation method with $p < 0.005$.
The median improvement in MSE over complete 3 s trajectories is 3.6\%. Additionally, the maximum percent improvement in average error magnitude for a timestep is 35\% at $0.93$s.

\subsection{ASLIP Hopper System}

Finally, we demonstrate the efficacy of this method on a more complicated robot model, the ASLIP (asymmetric spring loaded inverted pendulum) hopper \cite{poulakakis2009slip}, which is shown on the right in Figure \ref{fig:kalman_filtering_error_comparison_ASLIP}.
The state for the ASLIP hopper is defined as:
\begin{align}
    q = [x_b,y_b,\theta_b,x_t,y_t,\dot{x}_b,\dot{y}_b,\dot{\theta}_b]^T
\end{align}
In the aerial phase, the system uses ballistic dynamics with a fixed leg length and angle.
We used Lagrangian dynamics to generate the dynamic equations for the contact mode.
The guard for touchdown is based on the height of the toe reaching the ground, which is uncertain.
The guard for liftoff is based on the leg returning to rest length. The complete derivation of these dynamics is given in \cite{kong2021salt}.

For this example, 1000 trials were run for 5 seconds, which is approximately 4 hops depending on the sampled conditions, with system parameters:
\begin{align}
m_b = 1\mathrm{kg}, a_g = 9.8\mathrm{m/s^2}, l_b = 0.5\mathrm{m}, I_b = 1\mathrm{kg m^2}, \\
\nonumber k_h = 100\mathrm{nm/rad}, k_l = 100\mathrm{n/m}, l_0 = 1\mathrm{m}, \phi_0 = 0\mathrm{rad}
\end{align}
The initial state of the system is $y_b = 2.5\mathrm{m}, x_b = 0\mathrm{m}, \theta_b = 0\mathrm{rad}, x_t = 0\mathrm{m}, y_t = 1\mathrm{m}$ with zero initial velocity. The covariance of the initial positions was $10^{-6} I_5$ and the velocity covariance was $10^{-6} I_3$. 
The process noise is $W = 0.001I_8$ and the system measures all positions with covariance $V = 0.01I_5$. The height of the ground is uncertain with zero mean and a standard deviation of $0.01$m.

\addtolength{\textheight}{-3cm}   

Results for this system can be seen in Fig.~\ref{fig:kalman_filtering_error_comparison_ASLIP}. 
The uncertainty aware method primarily improved results in the $y$ direction for body position and velocity and toe position, as the guard saltation matrix primarily operates in the vertical direction due to the ground normal being vertical, but it propagates variations to other states through the dynamics outside of the touchdown event.
The uncertainty aware method outperforms the standard saltation method with $p < 0.005$.
The median improvement in MSE over complete five second trajectories is 54.7\%. Additionally, the maximum percent improvement in average error magnitude for a timestep is 60.1\% at 1.9s.

\section{Conclusion and Future Work}
In this work we derive a first order propagation law for guard and reset uncertainty by using the guard saltation matrix and a Jacobian of the reset map.
This uncertainty aware method outperforms the standard saltation matrix method at estimating resulting probability distributions simulated through uncertain hybrid guards and reset maps.
We then use this propagation law in the uncertainty aware SKF and achieved lower estimation error than the standard SKF.
The uncertainty aware SKF reduces the average estimation error overall, with up to 24-60\% improvement after impact events.

While these results aid in updating the covariance once a transition is believed to have occurred, there is still the problem of determining whether mode transitions have occurred.
Future work will extend these results to include explicit reasoning about which mode the system is in. 
One potential approach for reasoning about whether mode transitions have occurred is to modify multiple model estimation methods \cite{blom1988interacting,barhoumi2012observer,BALLUCHI2013915} to account for variable transition probabilities based on the estimated position relative to guard locations.

Additionally, this work handles single mode transitions and multiple mode transitions with known mode sequences. 
This can be extended to reasoning about simultaneous (or near simultaneous) contact utilizing the Bouligand derivative \cite{burden2016event,scholtes2012introduction}, which encodes multiple potential contact sequences.






\bibliographystyle{IEEEtran}
\bibliography{references}

\begin{thebibliography}{10}
\providecommand{\url}[1]{#1}
\csname url@samestyle\endcsname
\providecommand{\newblock}{\relax}
\providecommand{\bibinfo}[2]{#2}
\providecommand{\BIBentrySTDinterwordspacing}{\spaceskip=0pt\relax}
\providecommand{\BIBentryALTinterwordstretchfactor}{4}
\providecommand{\BIBentryALTinterwordspacing}{\spaceskip=\fontdimen2\font plus
\BIBentryALTinterwordstretchfactor\fontdimen3\font minus
  \fontdimen4\font\relax}
\providecommand{\BIBforeignlanguage}[2]{{%
\expandafter\ifx\csname l@#1\endcsname\relax
\typeout{** WARNING: IEEEtran.bst: No hyphenation pattern has been}%
\typeout{** loaded for the language `#1'. Using the pattern for}%
\typeout{** the default language instead.}%
\else
\language=\csname l@#1\endcsname
\fi
#2}}
\providecommand{\BIBdecl}{\relax}
\BIBdecl

\bibitem{mitcheetahstateestimation2018}
G.~{Bledt}, P.~M. {Wensing}, S.~{Ingersoll}, and S.~{Kim}, ``Contact model
  fusion for event-based locomotion in unstructured terrains,'' in \emph{IEEE
  International Conference on Robotics and Automation}, 2018, pp. 4399--4406.

\bibitem{bloesch2013state}
M.~Bloesch, M.~Hutter, M.~A. Hoepflinger, S.~Leutenegger, C.~Gehring, C.~D.
  Remy, and R.~Siegwart, ``State estimation for legged robots-consistent fusion
  of leg kinematics and {IMU},'' in \emph{Robotics: Science and Systems}, 2012,
  pp. 17--24.

\bibitem{hartley2020contact}
R.~Hartley, M.~Ghaffari, R.~M. Eustice, and J.~W. Grizzle, ``Contact-aided
  invariant extended {Kalman} filtering for robot state estimation,'' \emph{The
  International Journal of Robotics Research}, vol.~39, no.~4, pp. 402--430,
  2020.

\bibitem{varin2018constrained}
P.~Varin and S.~Kuindersma, ``A constrained {Kalman} filter for rigid body
  systems with frictional contact,'' in \emph{Workshop on the Algorithmic
  Foundations of Robotics}, 2018.

\bibitem{Back_Guckenheimer_Myers_1993}
A.~Back, J.~M. Guckenheimer, and M.~Myers, ``A dynamical simulation facility
  for hybrid systems,'' in \emph{Hybrid Systems}, ser. Lecture Notes in
  Computer Science.\hskip 1em plus 0.5em minus 0.4em\relax Springer Berlin /
  Heidelberg, 1993, vol. 736, pp. 255--267.

\bibitem{goebel2009hybrid}
R.~Goebel, R.~G. Sanfelice, and A.~R. Teel, ``Hybrid dynamical systems,''
  \emph{IEEE Control Systems Magazine}, vol.~29, no.~2, pp. 28--93, 2009.

\bibitem{LygerosJohansson2003}
J.~Lygeros, K.~H. Johansson, S.~N. Simic, J.~Zhang, and S.~S. Sastry,
  ``{Dynamical properties of hybrid automata},'' \emph{IEEE Transactions on
  Automatic Control}, vol.~48, no.~1, pp. 2--17, 2003.

\bibitem{kong2021salt}
N.~J. Kong, J.~J. Payne, G.~Council, and A.~M. Johnson, ``The {Salted} {Kalman}
  {Filter}: {Kalman} filtering on hybrid dynamical systems,''
  \emph{Automatica}, vol. 131, p. 109752, 2021.

\bibitem{GAO2021290}
Y.~Gao, C.~Yuan, and Y.~Gu, ``Invariant extended {Kalman} filtering for hybrid
  models of bipedal robot walking,'' \emph{IFAC-PapersOnLine}, vol.~54, no.~20,
  pp. 290--297, 2021, {Modeling, Estimation and Control Conference}.

\bibitem{blom1988interacting}
H.~A. Blom and Y.~Bar-Shalom, ``The interacting multiple model algorithm for
  systems with {Markovian} switching coefficients,'' \emph{IEEE Transactions on
  Automatic Control}, vol.~33, no.~8, pp. 780--783, 1988.

\bibitem{barhoumi2012observer}
N.~Barhoumi, F.~Msahli, M.~Djema{\"\i}, and K.~Busawon, ``Observer design for
  some classes of uniformly observable nonlinear hybrid systems,''
  \emph{Nonlinear Analysis: Hybrid Systems}, vol.~6, no.~4, pp. 917--929, 2012.

\bibitem{BALLUCHI2013915}
A.~Balluchi, L.~Benvenuti, M.~D.~D. Benedetto], and A.~Sangiovanni-Vincentelli,
  ``The design of dynamical observers for hybrid systems: Theory and
  application to an automotive control problem,'' \emph{Automatica}, vol.~49,
  no.~4, pp. 915 -- 925, 2013.

\bibitem{koutsoukos2002monitoring}
X.~Koutsoukos, J.~Kurien, and F.~Zhao, ``Monitoring and diagnosis of hybrid
  systems using particle filtering methods,'' in \emph{International Symposium
  on Mathematical Theory of Networks and Systems}, 2002.

\bibitem{koval2017manifold}
M.~C. Koval, N.~S. Pollard, and S.~S. Srinivasa, ``Pose estimation for planar
  contact manipulation with manifold particle filters,'' \emph{The
  International Journal of Robotics Research}, vol.~34, no.~7, pp. 922--945,
  2015.

\bibitem{AIZERMAN19581065}
M.~Aizerman and F.~Gantmakher, ``On the stability of periodic motions,''
  \emph{Journal of Applied Mathematics and Mechanics}, vol.~22, no.~6, pp.
  1065--1078, 1958.

\bibitem{hiskens2000trajectory}
I.~A. Hiskens and M.~Pai, ``Trajectory sensitivity analysis of hybrid
  systems,'' \emph{IEEE Transactions on Circuits and Systems I: Fundamental
  Theory and Applications}, vol.~47, no.~2, pp. 204--220, 2000.

\bibitem{leine2013dynamics}
R.~I. Leine and H.~Nijmeijer, \emph{Dynamics and bifurcations of non-smooth
  mechanical systems}.\hskip 1em plus 0.5em minus 0.4em\relax Springer Science
  \& Business Media, 2013, vol.~18.

\bibitem{burden2016event}
S.~A. Burden, S.~S. Sastry, D.~E. Koditschek, and S.~Revzen, ``Event--selected
  vector field discontinuities yield piecewise--differentiable flows,''
  \emph{SIAM Journal on Applied Dynamical Systems}, vol.~15, no.~2, pp.
  1227--1267, 2016.

\bibitem{biggio2014accurate}
M.~Biggio, F.~Bizzarri, A.~Brambilla, and M.~Storace, ``Accurate and efficient
  {PSD} computation in mixed-signal circuits: A time-domain approach,''
  \emph{IEEE Transactions on Circuits and Systems II: Express Briefs}, vol.~61,
  no.~11, pp. 905--909, 2014.

\bibitem{staunton2020noisy}
E.~J. Staunton and P.~T. Piiroinen, ``Discontinuity mappings for stochastic
  nonsmooth systems,'' \emph{Physica D: Nonlinear Phenomena}, vol. 406, p.
  132405, 2020.

\bibitem{johnson2016hybrid}
A.~M. Johnson, S.~A. Burden, and D.~E. Koditschek, ``A hybrid systems model for
  simple manipulation and self-manipulation systems,'' \emph{The International
  Journal of Robotics Research}, vol.~35, no.~11, pp. 1354--1392, 2016.

\bibitem{burden2018contraction}
S.~A. Burden, T.~Libby, and S.~D. Coogan, ``On contraction analysis for hybrid
  systems,'' 2018, arXiv:1811.03956.

\bibitem{KullbackLeiblerDiv}
S.~Kullback and R.~A. Leibler, ``{On Information and Sufficiency},'' \emph{The
  Annals of Mathematical Statistics}, vol.~22, no.~1, pp. 79 -- 86, 1951.

\bibitem{welch1995introduction}
G.~Welch and G.~Bishop, ``An introduction to the {Kalman} filter,'' University
  of North Carolina at Chapel Hill, Tech. Rep. 95--041, 1995, {U}pdated: July
  24, 2006.

\bibitem{dixon1946statistical}
W.~J. Dixon and A.~M. Mood, ``The statistical sign test,'' \emph{Journal of the
  American Statistical Association}, vol.~41, no. 236, pp. 557--566, 1946.

\bibitem{poulakakis2009slip}
I.~{Poulakakis} and J.~W. {Grizzle}, ``The spring loaded inverted pendulum as
  the hybrid zero dynamics of an asymmetric hopper,'' \emph{IEEE Transactions
  on Automatic Control}, vol.~54, no.~8, pp. 1779--1793, 2009.

\bibitem{scholtes2012introduction}
S.~Scholtes, \emph{Introduction to piecewise differentiable equations}.\hskip
  1em plus 0.5em minus 0.4em\relax Springer Science \& Business Media, 2012.

\end{thebibliography}
\end{document}